\pdfoutput=1

\documentclass[11pt]{article}

\usepackage[preprint]{acl}

\usepackage{times}
\usepackage{latexsym}

\usepackage[T1]{fontenc}
\usepackage[utf8]{inputenc}

\usepackage{microtype}

\usepackage{graphicx}

\usepackage{colortbl}
\usepackage{amssymb}
\usepackage{pifont}

\title{Unmask It! AI-Generated Product Review Detection in Dravidian Languages}

\author{
  \textbf{Somsubhra De} \and
  \textbf{Advait Vats}
\\
Indian Institute of Technology Madras\\
  \small{Correspondence: 
\texttt{somsubhra@outlook.in}
  }
}
\begin{document}
\maketitle
\begin{abstract}
The rise of Generative AI has led to a surge in AI-generated reviews, often posing a serious threat to the credibility of online platforms. Reviews serve as the primary source of information about products and services. Authentic reviews play a vital role in consumer decision-making. The presence of fabricated content misleads consumers, undermines trust and facilitates potential fraud in digital marketplaces. This study focuses on detecting AI-generated product reviews in Tamil and Malayalam, two low-resource languages where research in this domain is relatively under-explored. We worked on a range of approaches - from traditional machine learning methods to advanced transformer-based models such as Indic-BERT, IndicSBERT, MuRIL, XLM-RoBERTa and Malayalam-BERT. Our findings highlight the effectiveness of leveraging the state-of-the-art transformers in accurately identifying AI-generated content, demonstrating the potential in enhancing the detection of fake reviews in low-resource language settings.
\end{abstract}

\section{Introduction}

In recent years, rapid advancements in artificial intelligence (AI) have significantly transformed various domains, including online content generation. Among these, the rise of AI-generated product reviews has become a major concern. These reviews, often hard to tell apart from human-written ones, threaten the trust and reliability of online platforms by influencing consumer opinions and disrupting market fairness. Since most consumers rely heavily on reviews before purchasing a product, it is essential that they differentiate between human-written and AI-generated reviews before coming to a decision. Investigations have identified apps with thousands of five-star ratings, many of which are convincingly crafted by AI. A 2023 analysis of around a million reviews revealed that 25\% of top apps in popular categories on Google Play and 17\% on the iOS App Store had suspicious reviews. Double Verify's Fraud Lab reported a threefold increase in apps with AI-powered fake reviews in 2024 compared to the same period in 2023 \cite{koetsier2024fake}. In response to the growing issue, companies like Amazon have stated that it is using advanced AI to detect inauthentic product reviews \cite{amazonarticle}.

While AI-powered review detection has advanced significantly for English, research in Dravidian languages remains limited. With a growing number of online shoppers relying on local-language reviews, there is a clear need for effective detection systems. The challenge is further amplified by the prevalence of code-mixed content such as Tamil written in Roman script, English words in Tamil script, intra-sentential switching, etc. - which is common in product reviews. Variations in linguistic features including syntax, morphology, lexicon, make the process more complex. Developing robust detection systems for Dravidian languages could help address these challenges \& better serve the Dravidian community. Such systems could serve as a valuable use case for integration into e-retail platforms, thereby improving transparency and trust in online marketplaces.

This study\footnote{The data \& codes are publicly available at \url{https://github.com/somsubhra04/dravlangtech_ai-gen-prod-rev}} contributes to the domain in the following aspects:
\begin{itemize}
\setlength{\itemsep}{0pt}
  \item We explore a range of ML, DL and SoTA transformer models to determine effective methods for detecting AI-generated reviews in the given dataset.
  \item We analyze \& provide insights into the strengths, drawbacks of each model \& perform a detailed error analysis.
\end{itemize}

\section{Related Work}
With LLMs like ChatGPT becoming commonplace, human and AI-generated texts are increasingly blending together in areas such as news, reviews, and social media, making it increasingly harder to distinguish between them as LLMs continue to improve. The study \cite{fraser2024detecting} examines the data collection process for datasets used in AI and human-generated text detection, referencing \cite{su-etal-2024-adapting, tum-nlp-IDMGSP-Galactica-TRAIN-CG}, pointing out that these are carefully curated and controlled rather than being organically sourced from online sources. The issue ties back to the fundamental challenge of the lack of a tool capable of definitively distinguishing between these two types of text. As a result, researchers have had to rely on pre 2020(before widespread adoption of LLMs) texts as human-labeled data and generate AI text themselves. The study further notes that AI text detection tools such as GPTZero, Originality.ai and CopyLeaks exist, but none of these provide a definitive solution at this point. Instead, the most reliable approach, recommended in the study, is to aggregate the various tools' result to obtain a reliable outcome. Given that this is the state of detection for English, the challenge is even greater for low-resource languages like Tamil and Malayalam, where even fewer datasets are available for research.

While quite a few studies have focused on NLP in Dravidian languages \cite{dravidianlangtech-2023-speech,dravidianlangtech-2024-speech}, the application to product reviews is still a relatively new area. There are few studies that specifically focus on human and AI-text detection in Tamil or Malayalam. A related study, with some similarity \cite{farsi-etal-2024-cuet-binary} focused on the detection of fake news in Malayalam, where Task-1 involved the binary classification of the news into original or fake category. It explored various models, ranging from ensemble methods to deep learning and transformer-based models. Although it was the transformer models that achieved the highest performance- MuRIL-BERT, Indic-SBERT, and XLM-R recorded the highest F1 score of 0.86. In contrast, the deep learning model performed least effectively.

\cite{singhal-bedi-2024-transformers-dravidianlangtech} used XLM-RoBERTa-large for a multi-class sentiment analysis of code-mixed Tamil, where it achieved an F1-score of 0.21. The study identified it as their best performing model. The model was then fine tuned for 20 epochs, with maximum sequence length of 512, using the Adam optimizer and cross entropy loss as the loss function. 

Similarly, while deep learning models had the least performance in \cite{farsi-etal-2024-cuet-binary}, \cite{he-etal-2017-yzu} showcased that a combination of BiLSTM-CNN gave a higher F1-score than individual DL models, when used on a dataset comprising of English tweets.

\section{Task \& Dataset Description}
This binary classification task aims to distinguish between two categories of reviews: AI-generated (\textit{AI}) and human-written (\textit{HUMAN}). The detailed distribution of the datasets provided by \cite{premjith-2025-overview-detectai} is presented in Table~\ref{dataset-distribution}. As observed, the label distribution is even, with no signs of class imbalance.
\begin{table}[ht]
  \centering
  \begin{tabular}{lllll}
    \hline
    \textbf{Language}  & \textbf{Dataset} & \multicolumn{2}{c}{\textbf{Classes}} & \textbf{Total}\\
    \hline
    \textbf{}  & \textbf{} & \textbf{AI} & \textbf{HUMAN} & \textbf{}\\
    \hline
    Tamil  & Train & 405 & 403 & 808\\
  & Test  & 48 & 52 & 100\\
    \hline
    Malayalam & Train & 400 & 400 & 800\\
  & Test  & 100 & 100 & 200\\
    \hline
  \end{tabular}
  \caption{\label{dataset-distribution}
    Dataset distribution for Tamil and Malayalam product reviews
  }
\end{table}
\section{Methodology}
\subsection{Pre-processing \& Feature Extraction for DL \& ML Approach}
The following steps were applied to the raw text: \textit{data cleaning} (HTML tags, punctuation, digits and extra whitespaces were removed using regular expressions), \textit{tokenization} (text was tokenized into words for word-level analysis), and label encoding (target labels were converted into numerical labels using LabelEncoder). For \textit{feature extraction}, \textbf{TF-IDF} (TfidfVectorizer transformed the text into numerical vectors representing the importance of words in the document, with up to 5000 features including unigrams and bigrams) and \textbf{Word2Vec} embeddings (a Word2Vec model trained on the text data generated 100-dimensional word vectors, and an average vector was computed for each document) were applied. These features were combined into a single matrix to provide a richer representation of the text data and \textit{scaled} using StandardScaler to normalize the values, enhancing the performance of scale-sensitive models.
\subsection{Model Training}
\subsubsection{Transformers}
We applied several pre-trained transformer models which include \textbf{Indic-BERT} \cite{kakwani-etal-2020-indicnlpsuite}, \textbf{IndicSBERT} \cite{deode2023l3cubeindicsbertsimpleapproachlearning}, \textbf{MuRIL} \cite{khanuja2021muril}, \textbf{XLM-RoBERTa} \cite{conneau-etal-2020-unsupervised} and \textbf{Malayalam-BERT} \cite{joshi2023l3cubehindbertdevbertpretrainedbert}. AI4Bharat's Indic-BERT (Bidirectional Encoder Representations from Transformers) is a multilingual transformer model pre-trained on 12 major Indic languages, designed to capture language-specific nuances. The Indic sentence BERT (IndicSBERT) is a simplified variant of BERT tailored for 10 Indian languages, optimized for sentence-level understanding. Google's MuRIL (Multilingual Representations for Indian Languages)-base-cased is a transformer model trained on a large corpus of text data from 17 Indian languages, enhancing both language understanding and contextual embeddings. XLM-R is a multilingual transformer model built on the RoBERTa architecture, designed for cross-lingual understanding. It is trained on a massive amount of data across 100 languages, making it highly effective for various multilingual NLP tasks. Malayalam-BERT is a pre-trained transformer model specifically fine-tuned for Malayalam.

For each model, the \textit{AutoTokenizer} from the Hugging Face\footnote{\url{https://huggingface.co}} library was used to tokenize the text data automatically based on the specific model's architecture. The maximum sequence length was set to 128, and a \textit{batch size of 16} was used. The models were fine-tuned on 80\% of the train set (with \textit{random state = 42}) for \textit{3 epochs} with a \textit{learning rate of \( 2e^{-5} \)} and weight decay of 0.01. We've used Google Colab free version T4 GPU for running the experiments. Weights \& Biases (wandb) was used for experiment tracking, logging metrics and visualizing model performance during training.
\subsubsection{DL Models}
\textbf{CNN+BiLSTM}: The Convolutional Neural Network (CNN) layer captures local patterns and n-grams with 128 filters and a kernel size of 5, applying ReLU activation to introduce non-linearity - this layer helps the model learn spatial features from the input text. The following MaxPooling1D layer reduces the dimensionality, helping to retain only the most significant features. The Bidirectional Long Short-Term Memory (BiLSTM) layer captures both forward and backward dependencies in the text, helping to understand word context. Dropout layers with a rate of 0.5 are applied after the LSTM and Dense layers for regularization. Finally, a \textit{GlobalAveragePooling1D} layer reduces the BiLSTM output to a fixed-size vector, which is then passed through a Dense layer with 64 units and ReLU activation. The output layer is a Dense layer with softmax activation to produce class probabilities for multi-class classification. The model was trained on 80\% of the train set using the \textit{Adam optimizer} (\textit{learning rate: \( 1e^{-3} \)}) and sparse categorical cross-entropy loss over \textit{15 epochs} for both the Tamil \& Malayalam datasets respectively with a \textit{batch size of 32}.
\subsubsection{Traditional Approaches}
We trained Support Vector Machine (SVM) using Grid Search with 5-fold cross-validation to find the optimal combination of the hyper-parameters (kernel types - \textit{`linear', `rbf', `poly', `sigmoid'}, regularization parameter C values: \textit{[0.1, 1, 10, 100]} \& kernel coefficient gamma - \textit{`scale', `auto'}). Also, Random Forest with 100 estimators was trained on the feature set. XGBoost classifier, a gradient boosting algorithm, was trained with 100 estimators and learning rate of 0.1. We then combined both RF and XGBoost using a VotingClassifier for a soft voting.
\section{Results}
\subsection{Quantitative Analysis}
The \textit{macro-avg. F1-score}\footnote{\url{https://scikit-learn.org/1.5/modules/generated/sklearn.metrics.f1_score.html}} is utilized as the primary metric to assess the overall effectiveness of the system. Table~\ref{results} presents a detailed comparison of the performance across all models and approaches evaluated in this study.

For Task-1 (Tamil), \textbf{IndicSBERT outperforms} all models with the highest F1-score (96\%). IndicSBERT builds on multilingual BERT by fine-tuning it for cross-lingual sentence representation learning. This simple yet effective approach without explicit cross-lingual training enhances its ability to capture linguistic properties across languages.

Indic-BERT, MuRIL and XLM-RoBERTa demonstrate strong results but slightly lower than IndicSBERT. The Random Forest and XGBoost ensemble approach shows a relatively promising result however struggles with the Malayalam dataset, which might hint at challenges with the complexity of Malayalam. CNN+BiLSTM performs decently but lags behind transformer models. Malayalam-BERT outperforms in the Malayalam task with an impressive 92\% F1-score. This improved version of BERT was fine-tuned by \cite{joshi2023l3cubehindbertdevbertpretrainedbert} on publicly available monolingual Malayalam datasets, as existing multilingual models did not perform as well on downstream tasks. IndicSBERT is again the top performer along with MuRIL.
%threshold point for cam-ready
Interestingly, the XLM-RoBERTa-base model achieved a perfect precision \& recall (1.0) for the human \& AI classes respectively, resulting in zero false positives for human class and zero false negatives for AI-generated text. This means the model accurately identified all AI-generated texts. Whenever a text was classified as human-written, the prediction was always correct. Such performance is typically seen in imbalanced datasets where the model tends to favor the dominant class. However, this was not the case here, as both the training \& test data samples had only two instances extra, from either class.
The model made errors in six specific cases (shown in fig~\ref{fig3}). In each case, the model misclassified human-written texts as AI-generated. This suggests that while the model could learn distinct patterns in AI-generated data, human written texts with their more diverse styles, might have been harder to categorize. The MuRIL model also demonstrated a similar trend during validation, where it produced comparable results.

Coming to the DL models and traditional methods, the results shown in table~\ref{results3} indicate that the SVM model did well on both datasets, with macro-F1 scores of 0.85 and 0.77 in validation. However, its performance dropped on the test set, indicating some overfitting. The CNN+BiLSTM model showed a contrasting trend, with a relatively lower validation F1 but a significant improvement on the Tamil test set. However, the model struggled with generalization in Malayalam. The ensemble classifier experienced a performance drop in Malayalam on the test set, while maintaining strong results in Tamil.
 
\begin{table*}[ht]
  \centering
  \begin{tabular}{l|llll|llll}
    \hline
   %{\textbf{Model}}  & \multicolumn{4}{c}\textbf{Task-1 (Tamil)} & \multicolumn{4}{c}\textbf{Task-2 (Malayalam)}\\
       {\centering \textbf{Model}}  & \multicolumn{4}{c|}{\centering \textbf{Task-1 (Tamil)}} & \multicolumn{4}{c}{\centering \textbf{Task-2 (Malayalam)}} \\
    \cline{2-9}
& \textbf{P} & \textbf{R} & \textbf{F1} & \textbf{Acc.} & \textbf{P} & \textbf{R} & \textbf{F1} & \textbf{Acc.}\\
    \hline
Indic-BERT & 0.93 & 0.93 & 0.93 & 0.93 & 0.86 & 0.85 & 0.85 & 0.85\\
%\rowcolor{green!30}
\cellcolor{green!30}IndicSBERT & 0.96 & 0.96 & \cellcolor{green!30}\textbf{\textcolor{blue}{\underline {0.96}}} & 0.96 & 0.92 & 0.92 & 0.91 & 0.92\\
MuRIL & 0.94 & 0.94 & 0.94 & 0.94 & 0.9 & 0.9 & 0.9 & 0.9\\
XLM-RoBERTa & 0.94 & 0.94 & 0.94 & 0.94 & 0.87 & 0.87 & 0.87 & 0.87\\
\cellcolor{green!30}Malayalam-BERT & \multicolumn{4}{c|}{\centering -} & 0.93 & 0.93 & \cellcolor{green!30}\textbf{\textcolor{blue}{\underline {0.92}}} & 0.93\\
CNN+BiLSTM & 0.89 & 0.89 & 0.89 & 0.89 & 0.69 & 0.6 & 0.55 & 0.6\\
SVM & 0.77 & 0.77 & 0.77 & 0.77 & 0.65 & 0.65 & 0.65 & 0.65\\
Ensemble (RF+XGBoost) & 0.9 & 0.9 & 0.9 & 0.9 & 0.6 & 0.59 & 0.59 & 0.59\\
  \hline
  \end{tabular}
  \caption{\label{results}
    Performance of various models on the test-set (macro-averaged Precision, Recall, F1 and Accuracy scores from the {\color{red}\textit{best run}} for each approach have been mentioned)
  }
\end{table*}
\subsection{Qualitative Analysis}
We analyzed the characteristics of AI-generated and human-written product reviews on the train \& test sets (Tables ~\ref{ai-human-comparison},~\ref{ai-human-comparison-2}), focusing on their linguistic differences, common patterns \& sources of misclassification. A clear difference between the two categories is their length and complexity. In Malayalam, AI-generated reviews have a lower average word count compared to human-written reviews. A similar trend is observed in sentence length where AI-generated reviews tend to have shorter and more direct sentences compared to human reviews. However, in Tamil, this pattern is reversed - AI-generated reviews are significantly longer. This suggests that AI-generated content in Tamil may be overly descriptive compared to human reviews, which are often brief and to the point.

Interestingly, AI-generated Malayalam reviews have higher lexical diversity than human-written ones. This suggests that AI-generated reviews may use a broader vocabulary or introduce uncommon words that are less typical in natural user reviews. The opposite is observed in Tamil, where human reviews show higher lexical diversity compared to AI-generated reviews. We analyzed the false positives and false negatives for both tasks. A key question arises: \textit{For common misclassifications, which models are performing better \& predicting correctly?} Figures  ~\ref{fig8}, ~\ref{fig9} show all reviews that were misclassified by more than one transformer model. \textcolor{green}{\checkmark} indicates that the label has been correctly predicted by the model, while \textcolor{red}{\ding{55}} denotes incorrect prediction. For eg., in the 6th Tamil review, only XLM successfully identifies the AI-generated content, while all other models fail in this case.

We conducted a brief survey where 19 individuals proficient in Tamil or Malayalam reviewed misclassified samples. They were asked to categorize each sample as AI-generated or human-made based solely on their judgment, without access to ground truth labels or using any translation tools. Respondents who answered the survey most accurately observed the following: For Tamil, AI-generated text often uses uncommon words in multiple sentences, with some original Tamil words that have transitioned to colloquial usage. For Malayalam, they identified grammatical errors, unusual word choices, incorrect word placement, tense errors, unrelated words and lack of sentence continuity as indicators of AI-generated text.

\section{Conclusion}
The performance of various transformer and DL models was examined for classifying product reviews into Human written and AI generated categories for low resource languages like Tamil and Malayalam. While the DL models performed somewhat promisingly when multiple models were combined, their performance still did not match the performance of transformer models, especially for Malayalam, highlighting their strong capability \& efficacy in handling complex linguistic features in the Dravidian space. Among the best performing models were-Indic-BERT, IndicSBERT, MuRIL and XLM-RoBERTa. In general, all the models achieved a lower F1-score on Malayalam samples than Tamil, with Malayalam-BERT producing the best results for Malayalam classification. IndicSBERT performed best on the Tamil samples and closely followed Malayalam-BERT for Malayalam samples, making it the \textit{most efficient model when evaluated across both languages}.

\subsection{Future Work}
Future work will focus on employing LLMs (few-shot, CoT prompting, exploring RAG), trying ensemble methods with transformer-based models \& expanding to other low-resource languages for cross-lingual transfer learning. Additionally, we plan to conduct experiments on larger and more diverse datasets as the current study was limited in scope. This will help reproduce and assess the real-world applicability of our models, ensuring their effectiveness at scale. Also, one critical concern is the risk of misclassifying human-written text as AI-generated, leading to false positives. This can have significant consequences such as unwarranted censorship \& questioning of genuine user feedback. Moreover, the ethical implications of AI-generated text detection need to be considered, particularly regarding privacy and bias. An important future direction will be ensuring that the detection systems developed are both accurate and fair, minimizing the chances of misclassification.
\section{Limitations}
The transformer models, pre-trained on corpora created for different tasks, may limit their performance on review detection. The lack of sufficient data in low-resource languages hampers effective fine-tuning for this specific task. Additionally, the dataset used was not code-mixed. Compute limitations restricted the ability to fine-tune transformers efficiently. Also, we could have analyzed the misclassified examples to check for any possible bias in the transformers, which might have given useful insights. Specifically, it would have been important to examine whether the model shows biases towards certain groups of reviews, such as favoring specific dialects or writing patterns/styles. However, due to our lack of proficiency in Tamil and Malayalam, we were unable to carry out an in-depth analysis.
\section*{Acknowledgments}
We are thankful to the Organizers of the \emph{Fifth Workshop on Speech, Vision, and Language Technologies for Dravidian Languages} at NAACL 2025, especially Dr. Premjith B. \& Dr. Bharathi Raja Chakravarthi for their prompt responses to our queries.

\bibliography{camready}

\begin{thebibliography}{16}
\providecommand{\natexlab}[1]{#1}

\bibitem[{Chakravarthi et~al.(2023)Chakravarthi, Priyadharshini, M, Thavareesan, and Sherly}]{dravidianlangtech-2023-speech}
Bharathi~R. Chakravarthi, Ruba Priyadharshini, Anand~Kumar M, Sajeetha Thavareesan, and Elizabeth Sherly, editors. 2023.
\newblock \href {https://aclanthology.org/2023.dravidianlangtech-1.0/} {\emph{Proceedings of the Third Workshop on Speech and Language Technologies for Dravidian Languages}}. INCOMA Ltd., Shoumen, Bulgaria, Varna, Bulgaria.

\bibitem[{Chakravarthi et~al.(2024)Chakravarthi, Priyadharshini, Madasamy, Thavareesan, Sherly, Nadarajan, and Ravikiran}]{dravidianlangtech-2024-speech}
Bharathi~Raja Chakravarthi, Ruba Priyadharshini, Anand~Kumar Madasamy, Sajeetha Thavareesan, Elizabeth Sherly, Rajeswari Nadarajan, and Manikandan Ravikiran, editors. 2024.
\newblock \href {https://aclanthology.org/2024.dravidianlangtech-1.0/} {\emph{Proceedings of the Fourth Workshop on Speech, Vision, and Language Technologies for Dravidian Languages}}. Association for Computational Linguistics, St. Julian's, Malta.

\bibitem[{Conneau et~al.(2020)Conneau, Khandelwal, Goyal, Chaudhary, Wenzek, Guzm{\'a}n, Grave, Ott, Zettlemoyer, and Stoyanov}]{conneau-etal-2020-unsupervised}
Alexis Conneau, Kartikay Khandelwal, Naman Goyal, Vishrav Chaudhary, Guillaume Wenzek, Francisco Guzm{\'a}n, Edouard Grave, Myle Ott, Luke Zettlemoyer, and Veselin Stoyanov. 2020.
\newblock \href {https://doi.org/10.18653/v1/2020.acl-main.747} {Unsupervised cross-lingual representation learning at scale}.
\newblock In \emph{Proceedings of the 58th Annual Meeting of the Association for Computational Linguistics}, pages 8440--8451, Online. Association for Computational Linguistics.

\bibitem[{Deode et~al.(2023)Deode, Gadre, Kajale, Joshi, and Joshi}]{deode2023l3cubeindicsbertsimpleapproachlearning}
Samruddhi Deode, Janhavi Gadre, Aditi Kajale, Ananya Joshi, and Raviraj Joshi. 2023.
\newblock \href {https://arxiv.org/abs/2304.11434} {L3cube-indicsbert: A simple approach for learning cross-lingual sentence representations using multilingual bert}.
\newblock \emph{Preprint}, arXiv:2304.11434.

\bibitem[{Economic-Times(2023)}]{amazonarticle}
Economic-Times. 2023.
\newblock \href {https://retail.economictimes.indiatimes.com/news/e-commerce/e-tailing/using-advanced-ai-to-spot-and-remove-fake-customer-reviews-amazon/105308125} {Using advanced ai to spot and remove fake customer reviews: Amazon}.
\newblock (Accessed: 22 January 2025).

\bibitem[{Farsi et~al.(2024)Farsi, Eusha, Islam, Ali~Taher, Hossain, Ahsan, Das, and Hoque}]{farsi-etal-2024-cuet-binary}
Salman Farsi, Asrarul Eusha, Ariful Islam, Hasan~Mesbaul Ali~Taher, Jawad Hossain, Shawly Ahsan, Avishek Das, and Mohammed~Moshiul Hoque. 2024.
\newblock \href {https://aclanthology.org/2024.dravidianlangtech-1.29/} {{CUET}{\_}{B}inary{\_}{H}ackers@{D}ravidian{L}ang{T}ech {EACL}2024: Fake news detection in {M}alayalam language leveraging fine-tuned {M}u{RIL} {BERT}}.
\newblock In \emph{Proceedings of the Fourth Workshop on Speech, Vision, and Language Technologies for Dravidian Languages}, pages 173--179, St. Julian's, Malta. Association for Computational Linguistics.

\bibitem[{Fraser et~al.(2024)Fraser, Dawkins, and Kiritchenko}]{fraser2024detecting}
Kathleen~C. Fraser, Hillary Dawkins, and Svetlana Kiritchenko. 2024.
\newblock \href {https://doi.org/10.48550/arXiv.2406.15583} {Detecting ai-generated text: Factors influencing detectability with current methods}.
\newblock \emph{Preprint}, arXiv:2406.15583.

\bibitem[{He et~al.(2017)He, Yu, Lai, and Liu}]{he-etal-2017-yzu}
Yuanye He, Liang-Chih Yu, K.~Robert Lai, and Weiyi Liu. 2017.
\newblock \href {https://doi.org/10.18653/v1/W17-5233} {{YZU}-{NLP} at {E}mo{I}nt-2017: Determining emotion intensity using a bi-directional {LSTM}-{CNN} model}.
\newblock In \emph{Proceedings of the 8th Workshop on Computational Approaches to Subjectivity, Sentiment and Social Media Analysis}, pages 238--242, Copenhagen, Denmark. Association for Computational Linguistics.

\bibitem[{Joshi(2023)}]{joshi2023l3cubehindbertdevbertpretrainedbert}
Raviraj Joshi. 2023.
\newblock \href {https://arxiv.org/abs/2211.11418} {L3cube-hindbert and devbert: Pre-trained bert transformer models for devanagari based hindi and marathi languages}.
\newblock \emph{Preprint}, arXiv:2211.11418.

\bibitem[{Kakwani et~al.(2020)Kakwani, Kunchukuttan, Golla, N.C., Bhattacharyya, Khapra, and Kumar}]{kakwani-etal-2020-indicnlpsuite}
Divyanshu Kakwani, Anoop Kunchukuttan, Satish Golla, Gokul N.C., Avik Bhattacharyya, Mitesh~M. Khapra, and Pratyush Kumar. 2020.
\newblock \href {https://doi.org/10.18653/v1/2020.findings-emnlp.445} {{I}ndic{NLPS}uite: Monolingual corpora, evaluation benchmarks and pre-trained multilingual language models for {I}ndian languages}.
\newblock In \emph{Findings of the Association for Computational Linguistics: EMNLP 2020}, pages 4948--4961, Online. Association for Computational Linguistics.

\bibitem[{Khanuja et~al.(2021)Khanuja, Bansal, Mehtani, Khosla, Dey, Gopalan, Margam, Aggarwal, Nagipogu, Dave, Gupta, Gali, Subramanian, and Talukdar}]{khanuja2021muril}
Simran Khanuja, Diksha Bansal, Sarvesh Mehtani, Savya Khosla, Atreyee Dey, Balaji Gopalan, Dilip~Kumar Margam, Pooja Aggarwal, Rajiv~Teja Nagipogu, Shachi Dave, Shruti Gupta, Subhash Chandra~Bose Gali, Vish Subramanian, and Partha Talukdar. 2021.
\newblock \href {https://arxiv.org/abs/2103.10730} {Muril: Multilingual representations for indian languages}.
\newblock \emph{Preprint}, arXiv:2103.10730.

\bibitem[{Koetsier(2024)}]{koetsier2024fake}
J.~Koetsier. 2024.
\newblock \href {https://www.forbes.com/sites/johnkoetsier/2024/08/31/fake-ai-generated-reviews-flooding-app-stores/} {Fake ai-generated reviews flooding app stores}.
\newblock (Accessed: 22 January 2025).

\bibitem[{Premjith et~al.(2025)Premjith, K, Chakravarthi, Durairaj, Palani, Thavareesan, and Kumaresan}]{premjith-2025-overview-detectai}
B~Premjith, Nandhini K, Bharathi~Raja Chakravarthi, Thenmozhi Durairaj, Balasubramanian Palani, Sajeetha Thavareesan, and Prasanna~Kumar Kumaresan. 2025.
\newblock {O}verview of the {S}hared {T}ask on {D}etecting {AI} {G}enerated {P}roduct {R}eviews in {D}ravidian {L}anguages: {D}ravidian{L}ang{T}ech@{NAACL} 2025.
\newblock In \emph{Proceedings of the Fifth Workshop on Speech, Vision, and Language Technologies for Dravidian Languages}. Association for Computational Linguistics.

\bibitem[{Singhal and Bedi(2024)}]{singhal-bedi-2024-transformers-dravidianlangtech}
Kriti Singhal and Jatin Bedi. 2024.
\newblock \href {https://aclanthology.org/2024.dravidianlangtech-1.25/} {Transformers@{D}ravidian{L}ang{T}ech-{EACL}2024: Sentiment analysis of code-mixed {T}amil using {R}o{BERT}a}.
\newblock In \emph{Proceedings of the Fourth Workshop on Speech, Vision, and Language Technologies for Dravidian Languages}, pages 151--155, St. Julian's, Malta. Association for Computational Linguistics.

\bibitem[{Su et~al.(2024)Su, Cardie, and Nakov}]{su-etal-2024-adapting}
Jinyan Su, Claire Cardie, and Preslav Nakov. 2024.
\newblock \href {https://doi.org/10.18653/v1/2024.findings-naacl.95} {Adapting fake news detection to the era of large language models}.
\newblock In \emph{Findings of the Association for Computational Linguistics: NAACL 2024}, pages 1473--1490, Mexico City, Mexico. Association for Computational Linguistics.

\bibitem[{TUM(2023)}]{tum-nlp-IDMGSP-Galactica-TRAIN-CG}
TUM. 2023.
\newblock \href {https://huggingface.co/tum-nlp/IDMGSP-Galactica-TRAIN-CG} {Idmgsp-galactica-train-cg: A fine-tuned galactica model to detect machine-generated scientific papers}.
\newblock Accessed: 2025-01-29.

\end{thebibliography}
\appendix

\section{Appendix}
\label{sec:appendix}

\begin{table}[ht]
  \centering
  \footnotesize
  \begin{tabular}{l@{\hskip 5pt}l@{\hskip 5pt}l@{\hskip 6pt}l} 
    \hline
    \textbf{Language}  & \textbf{Metric} & \textbf{AI} & \textbf{Human}\\
    \hline
    \textbf{Tamil}  & Average Word Count & 7.904 & 5.700\\
                     & Average Sentence Length & 1.000 & 1.025\\
                     & Lexical Diversity & 0.992 & 0.987\\
    \hline
    \textbf{Malayalam} & Average Word Count & 12.155 & 16.815\\
                        & Average Sentence Length & 1.043 & 1.458\\
                        & Lexical Diversity & 0.983 & 0.939\\
    \hline
  \end{tabular}
  \caption{\label{ai-human-comparison} Analysis of AI-generated and human-written reviews for Tamil and Malayalam train-sets}
\end{table}

\begin{table}[ht]
  \centering
  \footnotesize
  \begin{tabular}{l@{\hskip 5pt}l@{\hskip 5pt}l@{\hskip 6pt}l} 
    \hline
    \textbf{Language}  & \textbf{Metric} & \textbf{AI} & \textbf{Human}\\
    \hline
    \textbf{Tamil}  & Average Word Count & 23.146 & 4.115\\
                     & Average Sentence Length & 2.333 & 1.019\\
                     & Lexical Diversity & 0.858 & 0.983\\
    \hline
    \textbf{Malayalam} & Average Word Count & 12.05 & 22.57\\
                        & Average Sentence Length & 1.00 & 1.73\\
                        & Lexical Diversity & 0.996 & 0.921\\
    \hline
  \end{tabular}
  \caption{\label{ai-human-comparison-2} Analysis of AI-generated and human-written reviews for Tamil and Malayalam test-sets}
\end{table}

\begin{figure*}[ht]
  \includegraphics[width=1.0\linewidth]{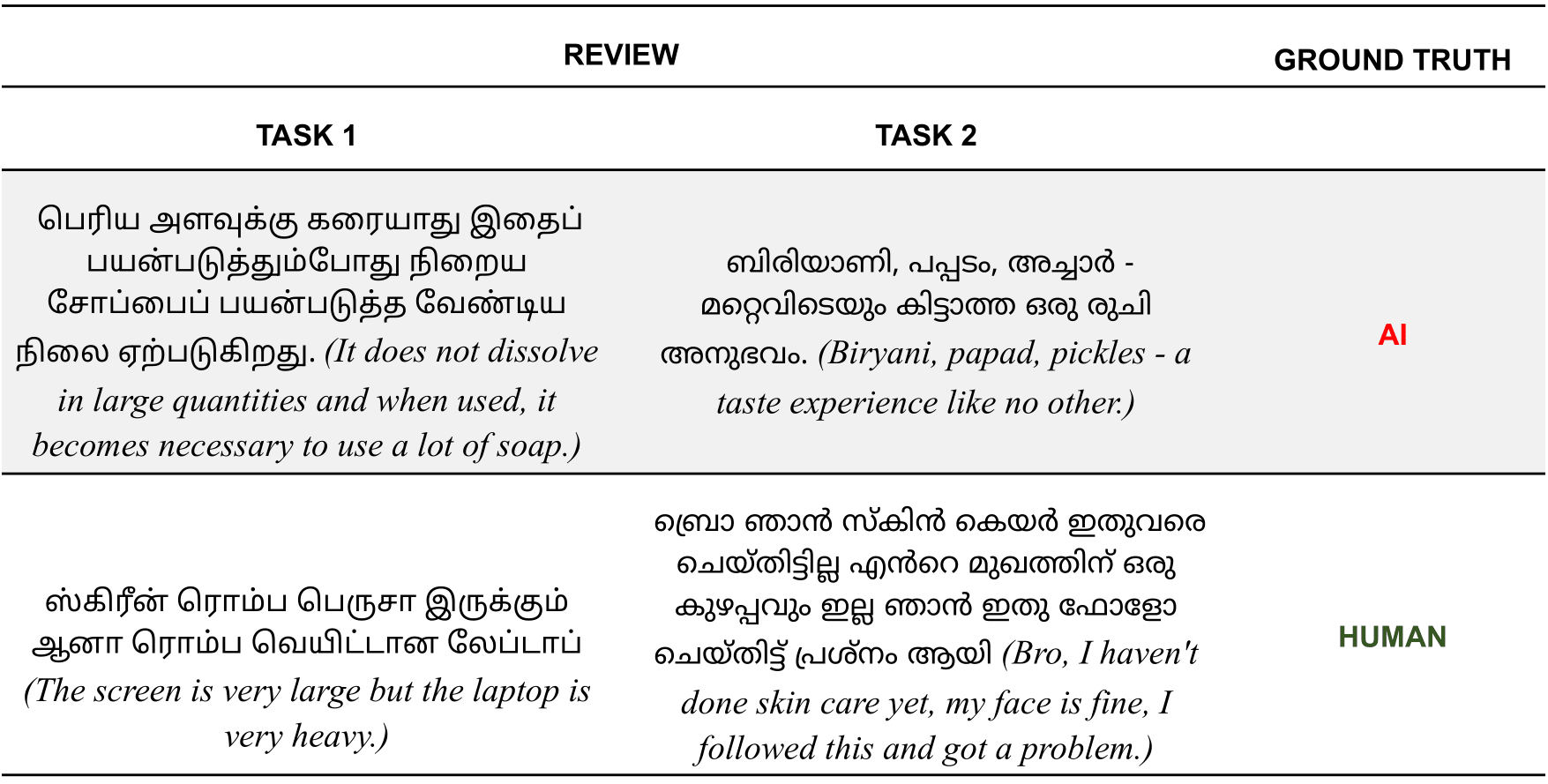}
  \caption {Sample Tamil and Malayalam Texts from the Training Set with English Translations* for Context\\\textit{(\textcolor{red}{Note*} The English translations may not fully capture the nuances, sentiment and cultural context inherent in the original Tamil and Malayalam texts. As a result, the English version might not reflect the true tone or intention.)}}
\end{figure*}

\begin{figure*}[ht]
  \includegraphics[width=0.48\linewidth]{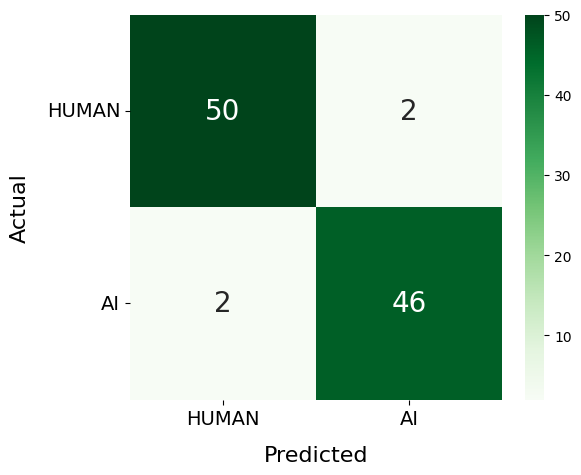} \hfill
  \includegraphics[width=0.48\linewidth]{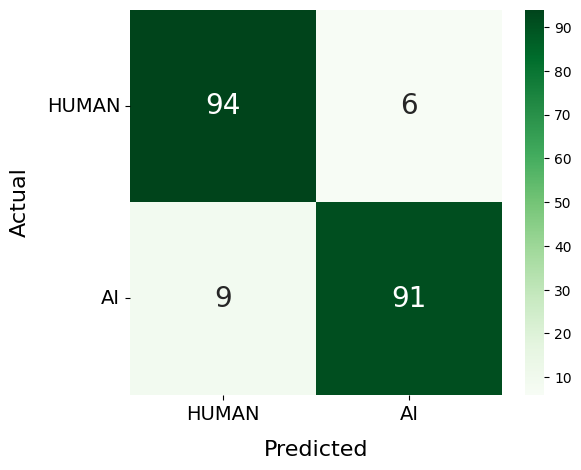}
  \caption {Confusion Matrix for the best runs on Tamil \& Malayalam test sets (using IndicSBERT \& Malayalam-BERT respectively)}
\end{figure*}

%\subsection{}
\begin{table*}[ht]
  \centering
  \renewcommand{\arraystretch}{1.2}
  \begin{tabular}{l|c|llll|llll}
    \hline
    {\textbf{Model}} & {\textbf{Category}} 
    & \multicolumn{4}{c|}{\textbf{Task-1 (Tamil)}} & \multicolumn{4}{c}{\textbf{Task-2 (Malayalam)}} \\
    \cline{3-10}
    &  & \textbf{P} & \textbf{R} & \textbf{F1} & \textbf{Acc.} & \textbf{P} & \textbf{R} & \textbf{F1} & \textbf{Acc.} \\
    \hline
    {Indic-BERT} 
    & \texttt{HUMAN} & 0.93 & 0.93 & 0.93 & - & 0.94 & 0.93 & 0.93 & - \\
    & \texttt{AI} & 0.94 & 0.94 & 0.94 & - & 0.94 & 0.93 & 0.93 & - \\
    & \textit{macro avg.} & 0.94 & 0.94 & 0.94 & 0.94 & 0.93 & 0.93 & 0.93 & 0.93 \\
    \hline
    {IndicSBERT} 
    & \texttt{HUMAN} & 0.97 & 0.99 & 0.98 & - & 0.95 & 0.95 & 0.95 & - \\
    & \texttt{AI} & 0.99 & 0.98 & 0.98 & - & 0.95 & 0.95 & 0.95 & - \\
    & \textit{macro avg.} & 0.98 & 0.98 & 0.98 & 0.98 & 0.95 & 0.95 & 0.95 & 0.95 \\
    \hline
    {MuRIL} 
    & \texttt{HUMAN} & 1 & 0.97 & 0.99 & - & 0.97 & 0.91 & 0.94 & - \\
    & \texttt{AI} & 0.98 & 1 & 0.99 & - & 0.92 & 0.97 & 0.95 & - \\
    & \textit{macro avg.} & 0.99 & 0.99 & \cellcolor{green!30}\textbf{0.99} & 0.99 & 0.95 & 0.94 & 0.94 & 0.94 \\
    \hline
    {XLM-RoBERTa} 
    & \texttt{HUMAN} & 1 & 0.86 & 0.92 & - & 1 & 0.84 & 0.91 & - \\
    & \texttt{AI} & 0.89 & 1 & 0.94 & - & 0.86 & 1 & 0.92 & -  \\
    & \textit{macro avg.} & 0.94 & 0.93 & 0.93 & 0.93 & 0.93 & 0.92 & 0.92 & 0.92 \\
    \hline
    {Malayalam-BERT} 
    & \texttt{HUMAN} & \multicolumn{4}{c|}{-} & 0.99 & 0.95 & 0.97 & - \\
    & \texttt{AI} & \multicolumn{4}{c|}{-} & 0.95 & 0.99 & 0.97 & - \\
    & \textit{macro avg.} & \multicolumn{4}{c|}{-} & 0.97 & 0.97 & \cellcolor{green!30}\textbf{0.97} & 0.97 \\
    \hline
  \end{tabular}
  \caption{\label{results2}
    Performance \textit{(Label-wise scores from the classification report)} of transformer-based models on the \textbf{validation set}
  }
\end{table*}

\begin{figure*}[ht]
  \centering
  \includegraphics[height=5cm]{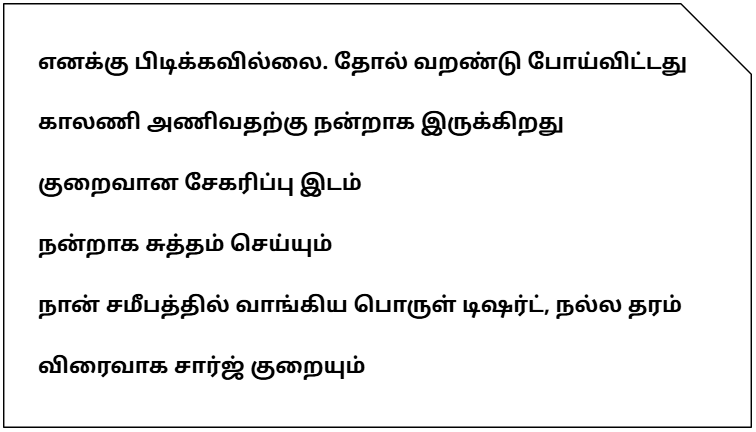} \hfill
  \includegraphics[height=5cm]{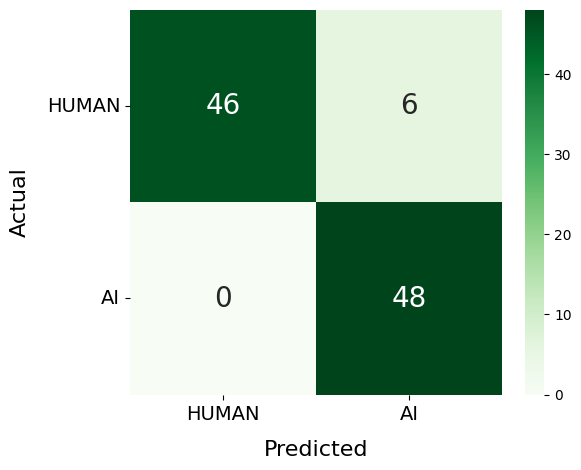}
  \caption{\label{fig3}Misclassified Tamil texts from the test set alongside the confusion matrix: XLM-R incorrectly predicted these 6 human-written reviews as AI-generated.}
\end{figure*}

\begin{table*}[ht]
  \centering
  \renewcommand{\arraystretch}{1.2}
  \begin{tabular}{l|c|llll|llll}
    \hline
    {\textbf{Model}} & {\textbf{Category}} 
    & \multicolumn{4}{c|}{\textbf{Task-1 (Tamil)}} & \multicolumn{4}{c}{\textbf{Task-2 (Malayalam)}} \\
    \cline{3-10}
    &  & \textbf{P} & \textbf{R} & \textbf{F1} & \textbf{Acc.} & \textbf{P} & \textbf{R} & \textbf{F1} & \textbf{Acc.} \\
    \hline
    {CNN+BiLSTM} 
    & \texttt{HUMAN} & 0.67 & 0.55 & 0.6 & - & 0.91 & 0.50 & 0.65 & - \\
    & \texttt{AI} & 0.66 & 0.76 & 0.7 & - & 0.66 & 0.95 & 0.78 & - \\
    & \textit{macro avg.} & 0.66 & 0.65 & 0.65 & 0.66 & 0.78 & 0.72 & 0.71 & 0.72 \\
    \hline
    {SVM} 
    & \texttt{HUMAN} & 0.78 & 0.95 & 0.86 & - & 0.81 & 0.7 & 0.75 & - \\
    & \texttt{AI} & 0.94 & 0.77 & 0.85 & - & 0.74 & 0.84 & 0.78 & - \\
    & \textit{macro avg.} & 0.86 & 0.86 & \cellcolor{blue!30}\textbf{0.85} & 0.85 & 0.77 & 0.77 & \cellcolor{blue!30}\textbf{0.77} & 0.77 \\
    \hline
{Ensemble (RF+XGBoost)} 
    & \texttt{HUMAN} & 0.78 & 0.8 & 0.79 & - & 0.72 & 0.65 & 0.68 & - \\
    & \texttt{AI} & 0.82 & 0.8 & 0.81 & - & 0.68 & 0.75 & 0.71 & - \\
    & \textit{macro avg.} & 0.8 & 0.8 & 0.8 & 0.8 & 0.7 & 0.7 & 0.7 & 0.7 \\
    \hline
  \end{tabular}
  \caption{\label{results3}
    Performance \textit{(Label-wise scores from the classification report)} of DL \& ML approaches on the \textbf{validation set}
  }
\end{table*}

\begin{figure*}[ht]
  \includegraphics[width=0.48\linewidth]{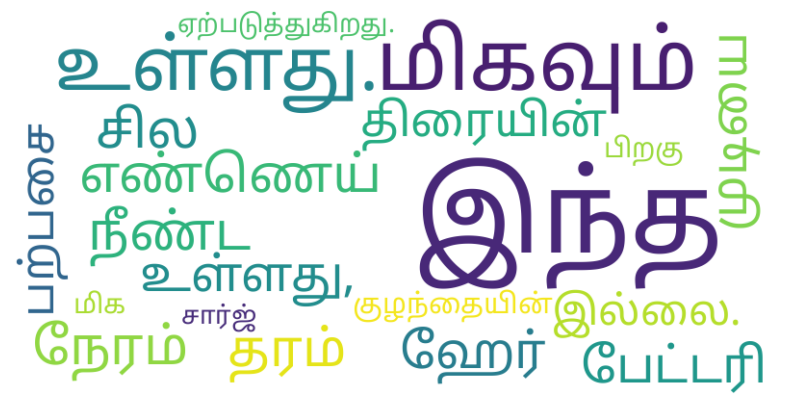} \hfill
  \includegraphics[width=0.48\linewidth]{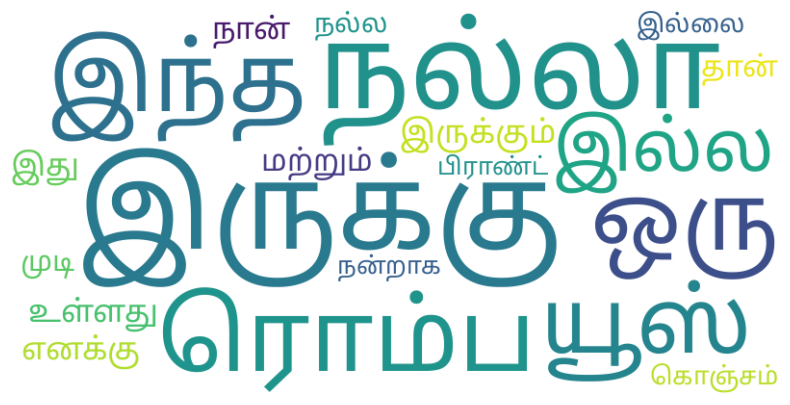}
  \caption {Most common words in AI-generated \textit{(left)} \& human-written \textit{(right)} reviews on the \textbf{Tamil train} set}
\end{figure*}

\begin{figure*}[ht]
  \includegraphics[width=0.48\linewidth]{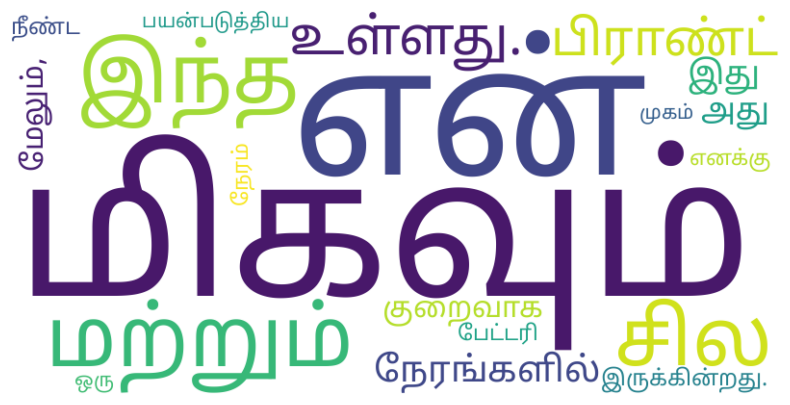} \hfill
  \includegraphics[width=0.48\linewidth]{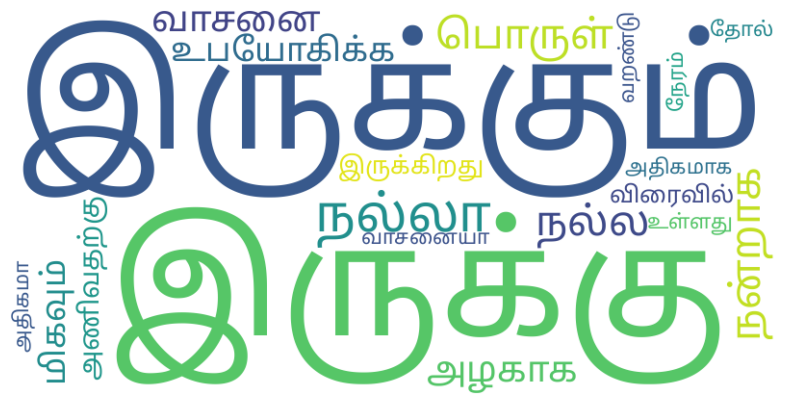}
  \caption {Most common words in AI-generated \textit{(left)} \& human-written \textit{(right)} reviews on the \textbf{Tamil test} set}
\end{figure*}

\begin{figure*}[ht]
  \includegraphics[width=0.48\linewidth]{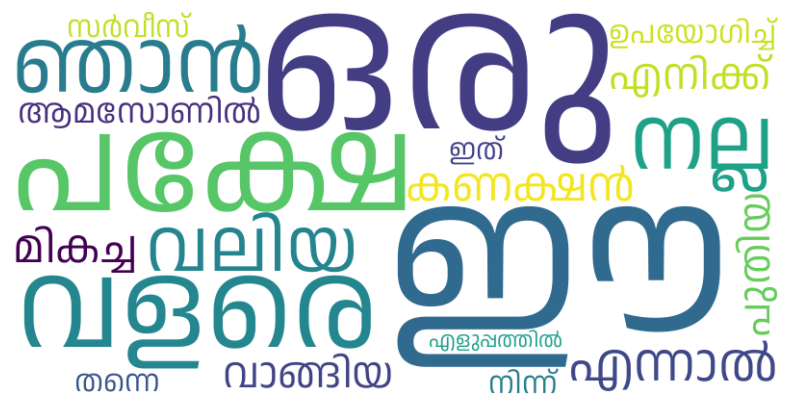} \hfill
  \includegraphics[width=0.48\linewidth]{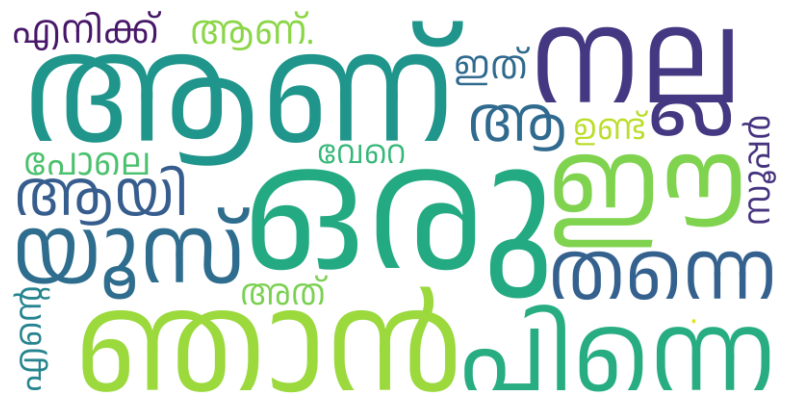}
  \caption {Most common words in AI-generated \textit{(left)} \& human-written \textit{(right)} reviews on the \textbf{Malayalam train} set}
\end{figure*}

\begin{figure*}[ht]
  \includegraphics[width=0.48\linewidth]{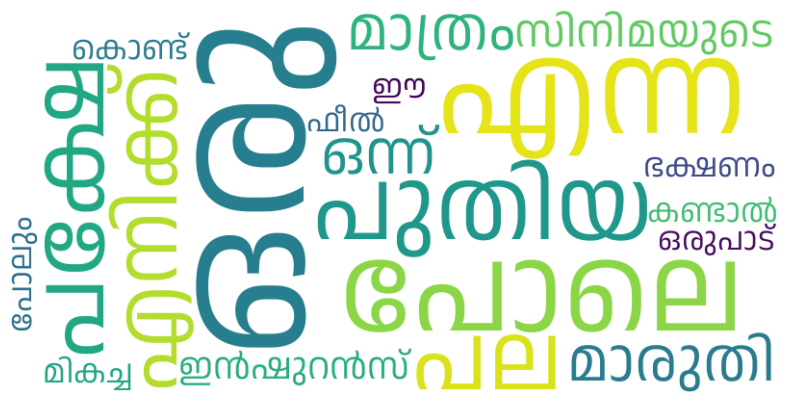} \hfill
  \includegraphics[width=0.48\linewidth]{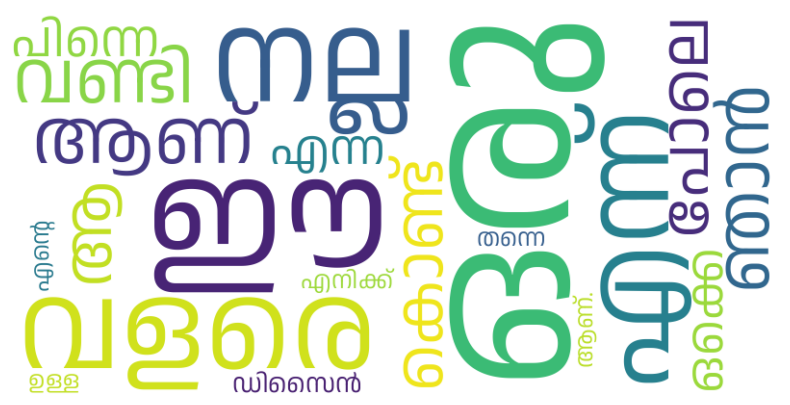}
  \caption {Most common words in AI-generated \textit{(left)} \& human-written \textit{(right)} reviews on the \textbf{Malayalam test} set}
\end{figure*}

\begin{figure*}[ht]
  \includegraphics[width=1.0\linewidth]{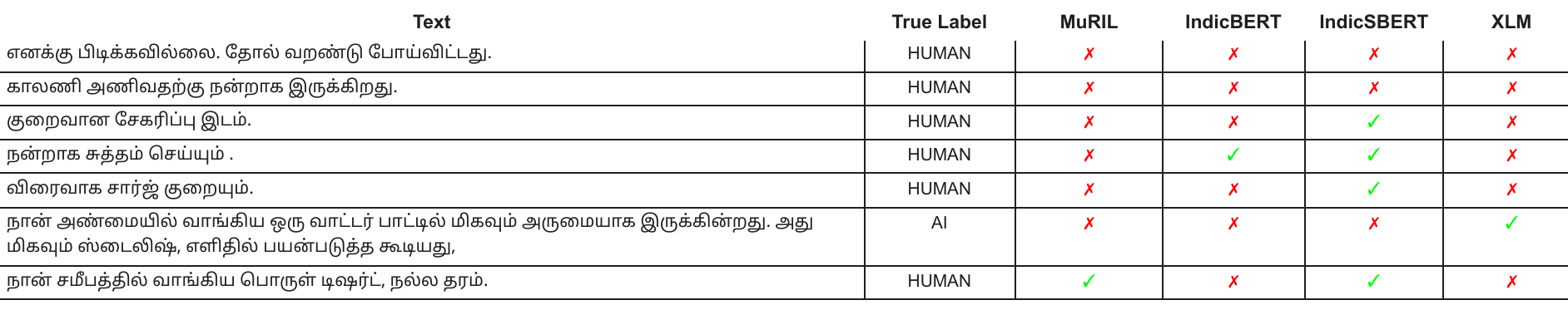}
  \caption {\label{fig8}Common misclassified reviews in Tamil}
\end{figure*}

\begin{figure*}[ht]
  \includegraphics[width=1.0\linewidth]{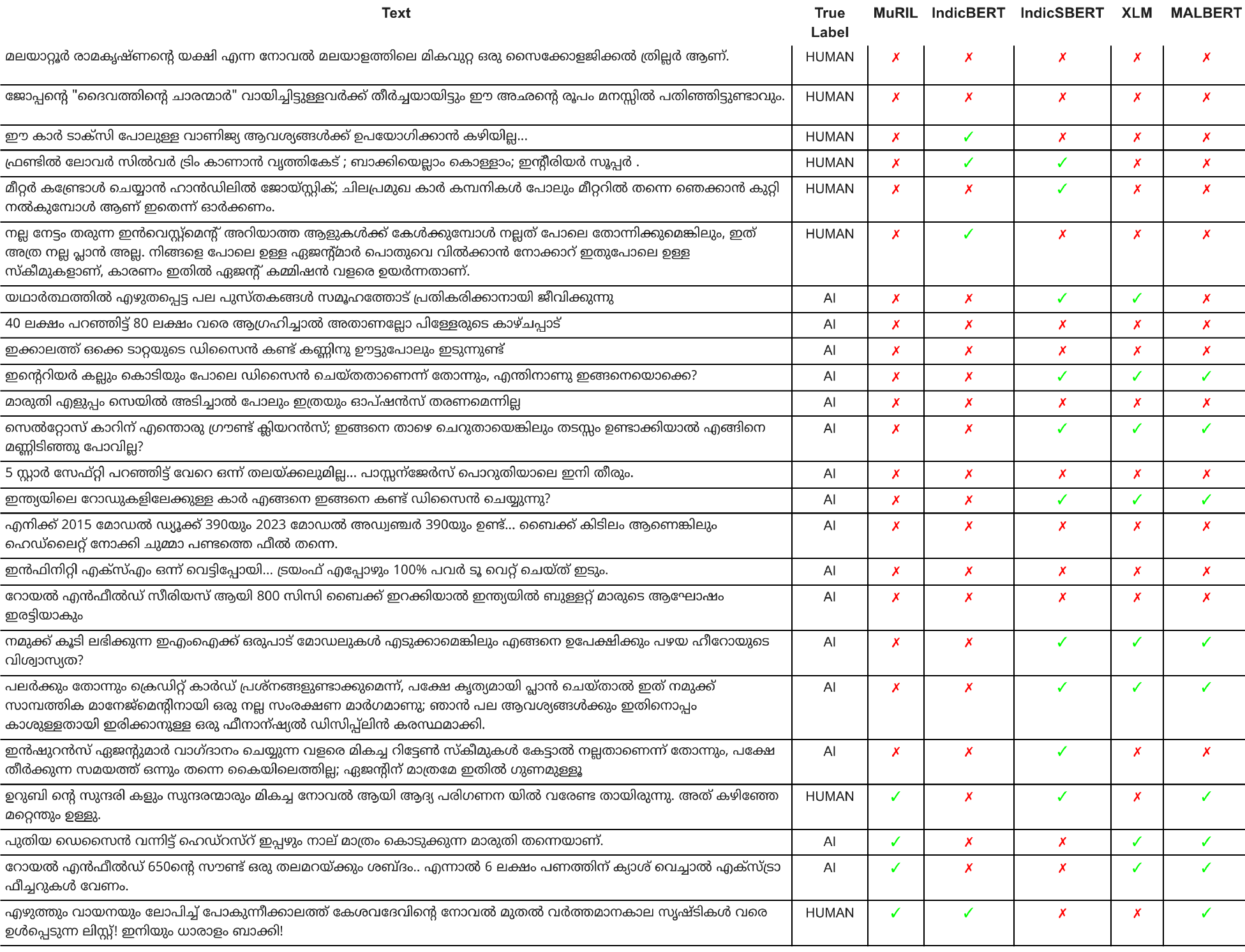}
  \caption {\label{fig9}Common misclassified reviews in Malayalam}
\end{figure*}

\end{document}